\def\year{2021}\relax
\documentclass[letterpaper]{article} 
\usepackage{aaai21}  
\usepackage{times}  
\usepackage{helvet} 
\usepackage{courier}  
\usepackage[hyphens]{url}  
\usepackage{graphicx} 
\usepackage{natbib}  
\usepackage{caption} 
\usepackage{amsfonts}
\usepackage{amsmath}
\usepackage{algorithm}
\usepackage{algorithmic}
\usepackage{multirow}
\usepackage{diagbox}
\usepackage{color} 
\makeatletter
\newcommand*\bigcdot{\mathpalette\bigcdot@{.5}}
\newcommand*\bigcdot@[2]{\mathbin{\vcenter{\hbox{\scalebox{#2}{$\m@th#1\bullet$}}}}}
\makeatother
\author{
  Chunnan Wang\textsuperscript{1}, Kaixin Zhang\textsuperscript{1}, Hongzhi Wang\textsuperscript{1,2}, Bozhou Chen\textsuperscript{1} \\
  \textsuperscript{1}Harbin Institute of Technology\\
  \textsuperscript{2}Peng Cheng Laboratory\\
  \texttt{\{WangChunnan, wangzh, bozhouchen\}@hit.edu.cn, 1170300216@stu.hit.edu.cn} \\
}

\title{Auto-STGCN: Autonomous Spatial-Temporal Graph Convolutional Network Search Based on Reinforcement Learning and Existing Research Results}
\begin{document}

\maketitle

\begin{abstract}
In recent years, many spatial-temporal graph convolutional network (STGCN) models are proposed to deal with the spatial-temporal network data forecasting problem. These STGCN models have their own advantages, i.e., each of them puts forward many effective operations and achieves good prediction results in the real applications. If users can effectively utilize and combine these excellent operations integrating the advantages of existing models, then they may obtain more effective STGCN models thus create greater value using existing work. However, they fail to do so due to the lack of domain knowledge, and there is lack of automated system to help users to achieve this goal. In this paper, we fill this gap and propose Auto-STGCN algorithm, which makes use of existing models to automatically explore high-performance STGCN model for specific scenarios. Specifically, we design Unified-STGCN framework, which summarizes the operations of existing architectures, and use parameters to control the usage and characteristic attributes of each operation, so as to realize the parameterized representation of the STGCN architecture and the reorganization and fusion of advantages. Then, we present Auto-STGCN, an optimization method based on reinforcement learning, to quickly search the parameter search space provided by Unified-STGCN, and generate optimal STGCN models automatically. Extensive experiments on real-world benchmark datasets show that our Auto-STGCN can find STGCN models superior to existing STGCN models with heuristic parameters, which demonstrates the effectiveness of our proposed method.
\end{abstract}

\section{Introduction}\label{section:1}

Spatial-temporal \underline{n}etwork \underline{d}ata \underline{f}orecasting (Spatial-Temporal NDF), which aims at predicting the future observations of a spatial-temporal network according to its historical series, is a fundamental research problem in spatial-temporal data mining. This problem has numerous real applications such as traffic speed forecasting~\cite{p18}, driver maneuver anticipation~\cite{p19} and human action recognition~\cite{p20}, and has attracted considerable research interest due to its importance. Researchers have proposed many methods to deal with it, and among them \underline{s}patial-\underline{t}emporal \underline{g}raph \underline{c}onvolutional \underline{n}etwork (STGCN) models are the most popular and effective solutions. STGCN models introduce \underline{g}raph \underline{c}onvolutional \underline{n}etwork (GCN)~\cite{p3,p4,p5,p6}, a powerful deep learning approach for graph-structured data, to learn high-level node representations~\cite{p7,p8,p9}, and combine GCN with other models or methods which are capable of modeling the temporal dependency, to extract high-quality spatial-temporal features directly from spatial-temporal network data. Compared with other solutions to the Spatial-Temporal NDF problem, which only take temporal information into account~\cite{p13,p22} or can only process standard grid structures rather than general domains~\cite{p10,p12,p21}, STGCN models can analyze the graph-structured time series more effectively, and thus make more accurate predictions~\cite{p15}. 												

Recently, many STGCN models~\cite{p14,p15,p16,p17} are proposed to deal with Spatial-Temporal NDF problems. We notice that these STGCN models have their own advantages, i.e., each of them puts forward many effective operations and achieves good prediction results in the real applications. If we can break the original combinations making excellent operations of different models capable of being combined together, then we can gain the following two benefits. (1) Obtaining novel and more powerful STGCN models. We may create more effective STGCN models by integrating the advantages of different models. (2) Promoting to realize the autonomous STGCN search. The effective operations provided by existing STGCN models constitute the search space of STGCN model, a key factor in achieving autonomous search of the STGCN model. We can utilize the effective optimization approach to explore this search space, and thus automatically design powerful STGCN models for non-experts according to their specific scenarios. 

However, achieving these benefits also brings two challenges. On the one hand, there is no unified framework to describe the design flow of STGCN models and lack of effective ways of representing various STGCN models. We need to fill this gap by figuring out the overall operation process of STGCN models, and thus provide guidance on how to collect operations from existing STGCN models and how to combine these operations. With the complete operation process and operation options, then we can find effective method of representing various STGCN models, thus realize automated STGCN search. On the other hand, there is lack of autonomous search methods designed for the STGCN model. We need design effective search method according to the characteristics of STGCN models, and thus quickly explore the huge search space of the STGCN model and discover powerful STGCN models.

In this paper, we overcome these challenges, and propose Auto-STGCN algorithm to make use of existing excellent models to automatically explore high-performance STGCN models. Specifically, we present Unified-STGCN framework, which reveals the overall operation process of STGCN models and summarizes operations of existing architectures. We use parameters to control the usage and characteristic attributes of each operation in Unified-STGCN, so as to realize the parameterized representation of the STGCN model and the reorganization and fusion of advantages. Then, we propose Auto-STGCN, an effective optimization method based on reinforcement learning, to quickly search the parameter search space provided by Unified-STGCN, and generate the optimal STGCN models automatically. Auto-STGCN considers both architecture-related parameters and training-related parameters during the optimization phase, and therefore can provide a complete solution, i.e., optimal STGCN structure combined with its optimal training setting, for the given Spatial-Temporal NDF problem.

Main contributions of our paper are concluded as follows:
\begin{itemize}
\item Unification: We unify various STGCN models under our Unified-STGCN framework, achieving the parameterized representation of the STGCN model. Unified-STGCN provides the parameter search space necessary for optimization method and deepens our understanding of popular STGCN models.
\item Automation: Our Auto-STGCN is an automated system for STGCN model development. It empowers non-experts to deploy STGCN models optimized for their specific scenarios. To the best of our knowledge, this is the first automated system in the field of Spatial-Temporal NDF.
\item Effectiveness: Extensive experiments on real-world benchmark datasets show that our Auto-STGCN can find STGCN models superior to existing STGCN models with heuristic parameters, which demonstrates the effectiveness of our proposed method.
\end{itemize}


\section{Prerequisite}\label{section:2}

In this section, we give the related concepts of Spatial-Temporal Network Data Forecasting (Section~\ref{section:2.1}), and introduce the state-of-the-art STGCN models (Section~\ref{section:2.2}). 

\subsection{Spatial-Temporal Network Data Forecasting}\label{section:2.1}

We firstly define spatial network and graph signal matrix, then describe Spatial-Temporal NDF problem using them.

\textbf{Definition 1: Spatial network $G$.} We use $G=(V,E,A)$ to denote the spatial information of a network, where $V$ is the set of vertices, $|V|=N$ denotes the number of vertices, $E$ denotes the set of edges, and $A\in \mathbb{R}^{N\times N}$ is the adjacency matrix of $G$. A spatial network $G$ can be either directed or undirected and its structure does not change with time. 

\textbf{Definition 2: Graph signal matrix $\chi_{G}^{t}$.} We use $\chi_{G}^{t}={(\chi_{G,v_{1}}^{t},\dots ,\chi_{G,v_{n}}^{t})}^{T}\in \mathbb{R}^{N\times C}$ to denote the observations of the spatial network $G=(V,E,A)$ at the time step $t$, where $C$ is the number of attribute features and $\chi_{G,v_{i}}^{t}$ denotes the values of all the features of node $v_{i}\in V$ at time step $t$. 

\textbf{Definition 3: Spatial-Temporal NDF Problem.} Given a spatial network $G$ and its historical graph signal matrices $\mathbb{X}=(\chi_{G}^{t-T+1}, \chi_{G}^{t-T+2}, \dots, \chi_{G}^{t})\in \mathbb{R}^{T\times N\times C}$, the Spatial-Temporal \underline{N}etwork \underline{D}ata \underline{F}orecasting (Spatial-Temporal NDF) problem aims at predicting the future observations of $G$: $\mathbb{Y}=(\chi_{G}^{t+1}, \chi_{G}^{t+2}, \dots, \chi_{G}^{t+T'})\in \mathbb{R}^{T'\times N\times C}$, where $T$ and $T’$ denote the length of the historical sequences and the target sequences to forecast respectively.

\begin{table*}[t]
\caption{Four stages that need to be gown through to get a STGCN model and its performance. Each stage needs to be well-designed so as to get a high-performance STGCN model.}
\vspace{-0.3cm}
\newcommand{\tabincell}[2]{\begin{tabular}{@{}#1@{}}#2\end{tabular}}
\centering
\resizebox{0.85\textwidth}{!}{
\smallskip\begin{tabular}{|m{3.1cm}|m{5.0cm}|m{3.2cm}|l|}
\hline
\textbf{Stage} & \textbf{Stage Function Description} & \textbf{Detail Contents} & \textbf{Solutions or Settings used in Existing Work} \\
\hline
Stage1: Input Transform Stage & Enhance representation power of the input historical sequence. & Input Structure (IS) & \tabincell{l}{$\bigcdot$ $IS_{1}$: Add dimension using fully-connected layer (\textit{Model4})\\$\bigcdot$ $IS_{2}$: None (\textit{Model1}, \textit{Model3}, \textit{Model2})} \\
\hline
\multirow{3}{*}{\tabincell{l}{\\ \\ \\ Stage2: Spatial-\\Temporal Embedding\\ Stage}} & Capture spatial-temporal dependencies from neighborhood or nearby times. & Spatial Information Processing Method (SIPM) & \tabincell{l}{$\bigcdot$ $SIPM_{1}$: Get new adjacent matrix using Pearson Coefficient (\textit{Model3})\\$\bigcdot$ $SIPM_{2}$: Spatial attention (\textit{Model2})\\$\bigcdot$ $SIPM_{3}$: Add mask weight to original adjacency matrix (\textit{Model4})\\
$\bigcdot$ $SIPM_{4}$: None (\textit{Model1})} \\
\cline{3-4}
& \multirow{2}{*}{\tabincell{l}{Existing work generally uses several\\ consecutive ST-blocks with same str-\\ucture to achieve this goal. The right\\ column shows the 3 main parts that\\ decide the structure of a ST-block.}} & Temporal Information Processing Method (TIPM) & \tabincell{l}{$\bigcdot$ $TIPM_{1}$: Temporal attention (\textit{Model2})\\$\bigcdot$ $TIPM_{2}$: Add learnable spatial and temporal embedding to the input series\\ (\textit{Model4})\\$\bigcdot$ $TIPM_{3}$: None (\textit{Model1}, \textit{Model3})}\\
\cline{3-4}
& & GCN-based Feature Embedding Structure (FES) & \tabincell{l}{$\bigcdot$ $FES_{1}$: TST-Sandwich Structure (\textit{Model1})\\$\bigcdot$ $FES_{2}$: GCN Layer (\textit{Model3})\\$\bigcdot$ $FES_{3}$: ST-Linear Structure (\textit{Model2})\\$\bigcdot$ $FES_{4}$: TS-Sliding Window Structure (\textit{Model4})}\\
\hline
Stage3: Output Transform Stage & Transform the output of the Stage2 (S2) into the expected prediction. & Output Structure (OS) & \tabincell{l}{$\bigcdot$ $OS_{1}$: LSTM Encoder Layer + LSTM Decoder Layer (\textit{Model3})\\$\bigcdot$ $OS_{2}$: One Fully Connected Layer (\textit{Model1}, \textit{Model2})\\$\bigcdot$ $OS_{3}$: Resize + Multi-Output Fully Connected Layer (\textit{Model4})} \\
\hline
\multirow{4}{*}{\tabincell{l}{\\ \\ \\ \\ Stage4: Training Stage}} & \multirow{4}{*}{\tabincell{l}{\\ \\ \\ Determine suitable training setting\\ for the designed STGCN model.}} & Loss Function (LF) & \tabincell{l}{$\bigcdot$ $LF_{1}$: MSE loss (\textit{Model1}, \textit{Model2}, \textit{Model3})\\$\bigcdot$ $LF_{2}$: Huber loss (\textit{Model4})}\\
\cline{3-4}
& & Batch Size (BS) & \tabincell{l}{$\bigcdot$ $BS_{1}$: 32 (\textit{Model3}, \textit{Model4})\\$\bigcdot$ $BS_{2}$: 50 (\textit{Model1})\\$\bigcdot$ $BS_{3}$: 64 (\textit{Model2})}\\
\cline{3-4}
& & Initial Learning Rate (ILR) & \tabincell{l}{$\bigcdot$ $ILR_{1}$: 1e-3 (\textit{Model1}, \textit{Model3}, \textit{Model4})\\$\bigcdot$ $ILR_{2}$: 7e-4 (\textit{Model3})\\$\bigcdot$ $ILR_{3}$: 1e-4 (\textit{Model3}, \textit{Model2})}\\
\cline{3-4}
& & Optimization Function (OF) & \tabincell{l}{$\bigcdot$ $OF_{1}$: RMSprop + StepDecay (decay rate of 0.7 every 10 epochs) (\textit{Model1})\\$\bigcdot$ $OF_{2}$: Adam (\textit{Model2}, \textit{Model3})\\$\bigcdot$ $OF_{3}$: Adam + PolyScheduler (\textit{Model4})}\\
\hline
\end{tabular}
}
\label{table1}
\vspace{-0.3cm}
\end{table*}

\subsection{Spatial-Temporal Graph Convolution Network Models}\label{section:2.2}

The key to solve Spatial-Temporal NDF problems is to capture spatial dependencies and temporal dependencies from spatial-temporal network data, and utilize these spatial-temporal features to make prediction. 

Recently, many Spatial-temporal graph convolutional network (STGCN) models are proposed to effectively deal with Spatial-Temporal NDF problems. They present various methods to capture dynamic spatial-temporal features of graph-structured time series. For example, \cite{p14} uses a GCN in spatial dimension to capture spatial dependencies from neighborhood and a gated CNN along temporal dimension to exploit temporal dependencies from nearby times. 
\cite{p15} constructs localized spatial-temporal graphs which connect individual spatial graphs of adjacent time steps into one graph, then synchronously captures localized spatial-temporal correlations in these localized spatial-temporal graphs using GCNs. \cite{p17} assumes that spatial correlations only depend on nodes with similar patterns and redefines connectivity of the graph according to the similarity among nodes. It utilizes GCN and the newly defined adjacency matrix to capture spatial correlations from most related regions, then uses multi-layer LSTM network to capture temporal relationships. 

Existing STGCN models have their own advantages, and have achieved extraordinary performance on many real applications~\cite{p18}. In this paper, we try to design more effective STGCN models by making use of valuable methods provided by them, achieving the fusion of advantages of existing STGCN models.

\section{Unified-STGCN: Unified STGCN Framework}\label{section:3}

We first propose our unified framework Unified-STGCN, and further explain how existing STGCN models fit in the framework (Section~\ref{section:3.1}). Based on these example models, we outline the shortcomings of existing STGCN models and give the parameterized representation of STGCN models on the basis of Unified-STGCN (Section~\ref{section:3.2}).

\subsection{Unified-STGCN}\label{section:3.1}

We analyze the workflow of existing STGCN models and sum up 4 stages of STGCN model design: Input Transform Stage, Spatial-Temporal Embedding Stage, Output Transform Stage and Training Stage. We define 9 parameters to describe the main content of these stages and thus realize the parameterized representation of the STGCN model. In addition, taking 4 state-of-the-art STGCN models, which are denoted by \textit{Model1}~\cite{p14}, \textit{Model2}~\cite{p16}, \textit{Model3}~\cite{p17} and \textit{Model4}~\cite{p15} respectively, as example, we provide the options for each parameter. Table~\ref{table1} summarizes the main contents of this part, and detailed introductions are as follows.

\begin{figure*}[t]
\centering
\includegraphics[width=0.75\textwidth]{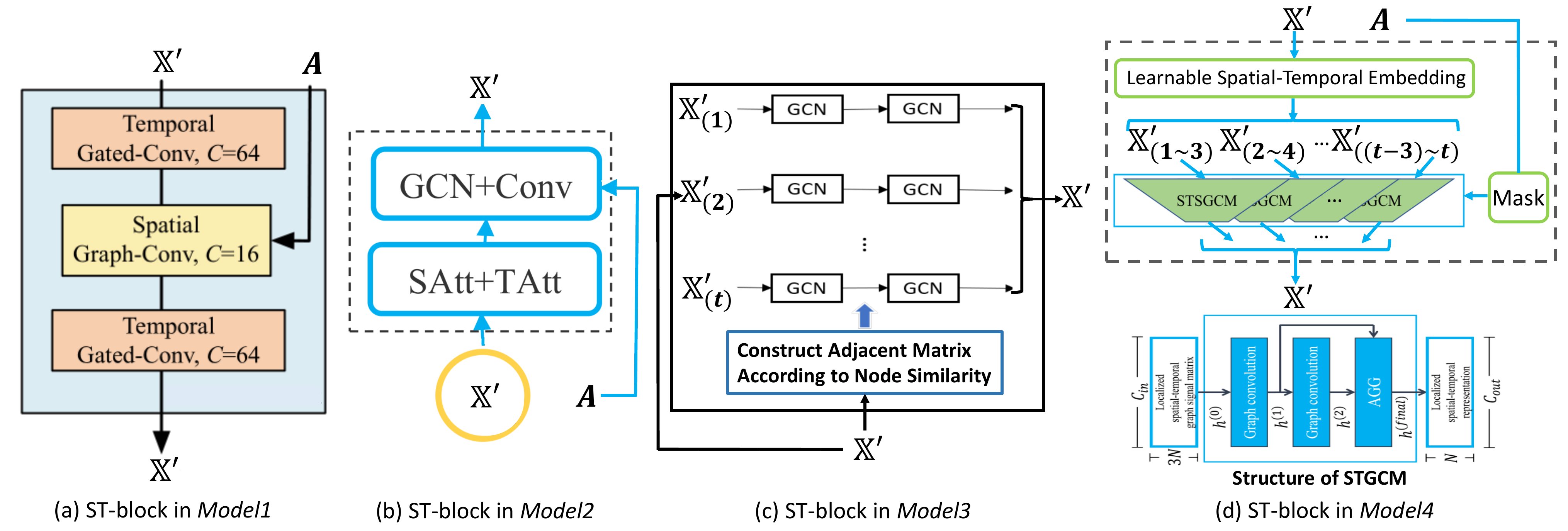}
\vspace{-0.3cm}
\caption{The structures of ST-blocks of existing STGCN models.}
\label{fig1}
\vspace{-0.55cm}
\end{figure*}

\subsubsection{Stage1: Input Transform Stage.}

The target of this stage is to improve the representation power of the input historical sequence $\mathbb{X}=(\chi_{G}^{t-T+1}, \chi_{G}^{t-T+2}, \dots, \chi_{G}^{t})\in \mathbb{R}^{T\times N\times C}$ of a Spatial-Temporal NDF problem. \textit{Model4} presents to use a fully connected layer at the top of the network to transform $\mathbb{X}$ into a high-dimension space so as to achieve this goal, whereas the other 3 models do not include this stage. In Unified-STGCN, we utilize parameter: Input Structure (\textit{IS}) to describe the detailed operation of this stage, and $IS_{1}$ and $IS_{2}$ denote the solution provided by \textit{Model4} and the other 3 models respectively.
\begin{equation}
\setlength{\abovedisplayskip}{0.05cm}
\setlength{\belowdisplayskip}{-0.00cm}
\begin{split}
&IS_1:  \mathbb{X}^{'}=FullyConnected(\mathbb{X})\in \mathbb{R}^{T\times N\times C^{'}}\\
&IS_2: \mathbb{X}^{'}=\mathbb{X}\in \mathbb{R}^{T\times N\times C}
\end{split}
\end{equation}

\subsubsection{Stage2: Spatial-Temporal Embedding Stage.}

This is the most important stage in STGCN model design. It aggregates high-level spatial-temporal correlations of the entire network series for final prediction. It takes the $\mathbb{X}^{'}$ provided by Stage1 and the input adjacency matrix $A$ of the Spatial-Temporal NDF problem as inputs, and outputs high-level representations of nodes in the network by encoding local graph structures and node attributes at different time steps.

Existing work generally uses several consecutive \underline{S}patial-\underline{T}emporal \underline{blocks} (ST-blocks) with same structure to achieve this goal. The structure of a ST-block is decided by 3 parameters: Spatial Information Processing Method (\textit{SIPM}), Temporal Information Processing Method (\textit{TIPM}) and GCN-based Feature Embedding Structure (\textit{FES}). A ST-block firstly adjusts input network series and adjacency matrix according to \textit{SIPM} and \textit{TIPM}, then utilizes \textit{FES} to process them and thus gets high-quality spatial-temporal features. Equation (2) describes its workflow. Since one ST-block can only capture low-level spatial-temporal features, existing models generally stack multiple ST-blocks to form deep models for more complicated features~\cite{p25}.
\begin{equation}
\setlength{\abovedisplayskip}{0.05cm}
\setlength{\belowdisplayskip}{-0.00cm}
\begin{split}
&A^{'}=SIPM(\mathbb{X}^{'},A),\ \ \mathbb{X}^{'}=TIPM(\mathbb{X}^{'})\\
&\mathbb{X}^{'}=FES(\mathbb{X}^{'},A^{'},A)\in \mathbb{R}^{T^{''}\times N\times FI} (T^{''}\leq T)
\end{split}
\end{equation}

We summarize the ST-blocks used in existing STGCN models, and find that their structures are quite different. \textit{Model1} skips \textit{SIPM} and \textit{TIPM} steps, and directly uses a TST-Sandwich Structure (as is shown in Figure~\ref{fig1} (a)), where GCN is used in the middle for extracting spatial features and two gated CNNs are applied for extracting temporal features, to jointly process graph-structured time series. \textit{Model2} introduces attention mechanism to adaptively capture dynamic correlations on the given network. It uses spatial attention to dynamically adjust impacting weights between nodes and uses temporal attention to dynamically adjust the input by merging relevant information, then feeds the adjusted input and matrix into convolution layers for getting high-quality features. 
As for \textit{Model3}, it assumes that spatial correlations only depend on the nodes with similar patterns~\cite{p24}. Its ST-block constructs a new adjacent matrix according to node similarity, then apply new matrix to GCN layers to extract high-level features. Figure~\ref{fig1} (c) gives the structure of this ST-block. The ST-block of \textit{Model4} equips position embedding to the input spatial-temporal network series so that each node contains time attributes, and adds a learnable mask matrix to the original adjacency matrix to adjust the aggregation weights so that the aggregation becomes more reasonable. Then, it utilizes a TS-Sliding Window Structure, which deploys multiple individual STSGCMs\footnote{STSGCM used in \textit{Model4} constructs localized spatial-temporal graphs which connect individual spatial graphs of adjacent time steps into one graph, then uses GCN operations to simultaneously capture localized spatial-temporal correlations.} on different time periods (as is shown in the top of Figure~\ref{fig1} (d)), to extract long-range spatial-temporal features. 

Existing ST-blocks have provided us with many effective methods for designing ST-blocks. We collect the options for SIPM, TIPM and FES parameters from state-of-the-art ST-blocks (as is shown in Table~\ref{table1}), and thus construct a powerful search space of the ST-block structure. Later, we will utilize this search space to realize the automatic design of ST-blocks, constructing more powerful STGCN models.

\begin{figure*}[t]
\centering
\includegraphics[width=0.75\textwidth]{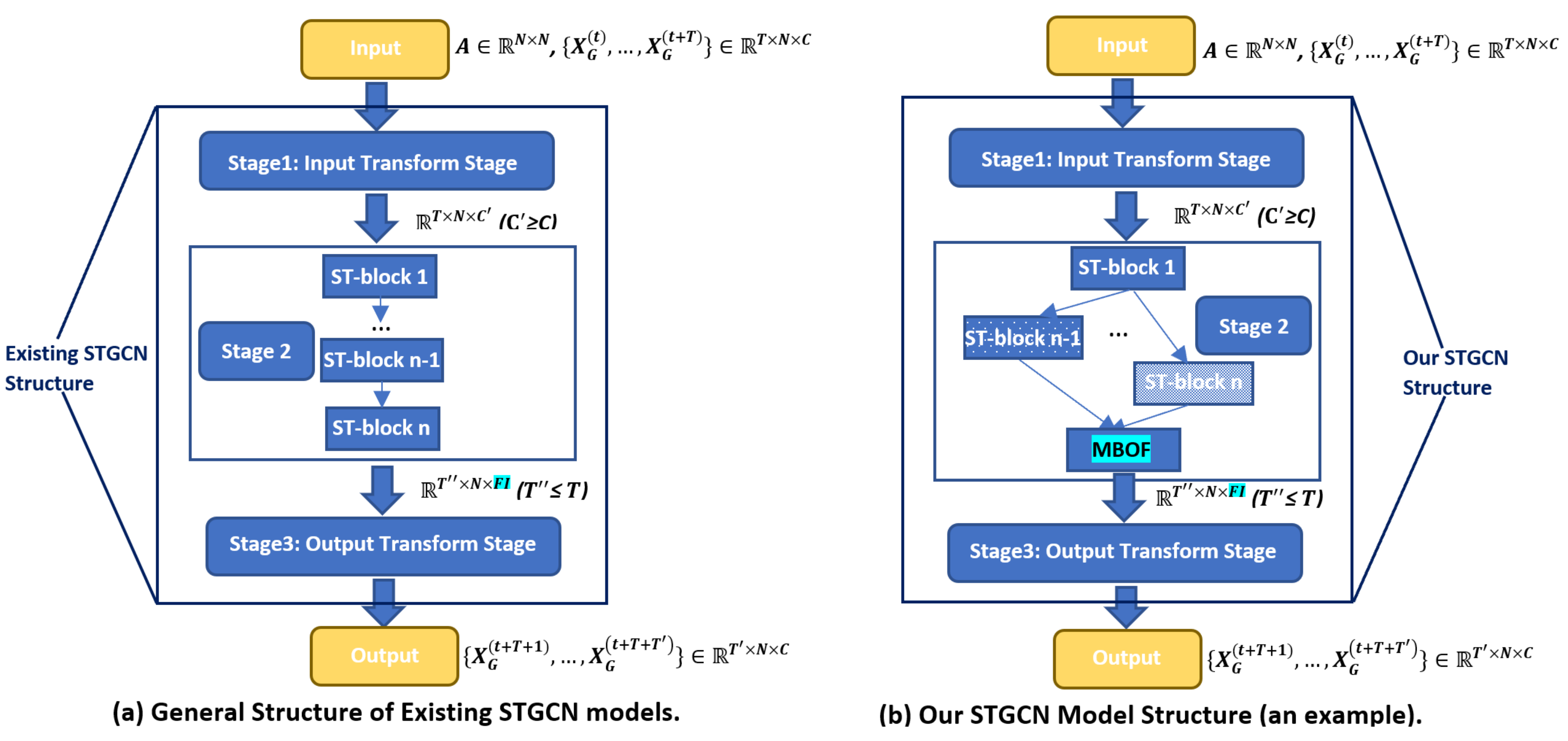}
\vspace{-0.3cm}
\caption{The structures of existing STGCN models and our STGCN models. The structures of existing STGCN models and our STGCN models. ST-blocks with different texture patterns have different structural settings.}
\label{fig2}
\vspace{-0.55cm}
\end{figure*}

\subsubsection{Stage3: Output Transform Stage.}

Stage3 is the last step of STGCN structure design. This stage aims at transforming the output of Stage2 into the expected prediction. Existing STGCN models propose many effective solutions. For example, 
\textit{Model1} and \textit{Model2} directly apply a fully connected layer to achieve the final transformation. \textit{Model3} utilizes a LSTM-based encoder-decoder method to generate the multi-step prediction. \textit{Model4} notices that there is heterogeneity in spatial-temporal data, i.e., each node may exhibit different properties at different time steps, and deploys multiple two-fully-connected layers to generate predictions of each time step to further improve the prediction performance. 

In Unified-STGCN, we utilize parameter: Output Structure (\textit{OS}) to describe the detailed operation of this stage, and $OS_{1}$, $OS_{2}$ and $OS_{3}$ denote the solution provided by \textit{Model3}, \textit{Model1} and \textit{Model2}, \textit{Model4} respectively.
\begin{equation}
\setlength{\abovedisplayskip}{0.1cm}
\setlength{\belowdisplayskip}{-0.02cm}
\mathbb{Y}=OS(\mathbb{X}^{'})\in \mathbb{R}^{T^{'}\times N\times C} 
\end{equation}

\subsubsection{Stage4: Training Stage.} 

The performance of a STGCN model depends not only on the network structure but also on the training setting. Therefore, finding a suitable training method for the STGCN model is also an important task, and we consider the model training as the fourth stage of STGCM model design. In Unified-STGCN, we summarize four parameters to describe the training setting of a STGCN model: Loss Function (\textit{LF}), Batch Size (\textit{BS}), Initial Learning Rate (\textit{ILR}) and Optimization Function (\textit{OF}). Existing works find out some training settings which are suitable for training STGCN models through repeated attempts. We collect the options for \textit{LF}, \textit{BS}, \textit{ILR} and \textit{OF} parameters from existing work (as is shown in Table~\ref{table1}), and thus construct a powerful search space of the STGCN model training. Later, we will use this search space to automatically design optimal training setting for our newly designed STGCN models.

\subsection{Parameterized Representation of STGCN}\label{section:3.2}

Nine parameters defined in Unified-STGCN clearly describe the design process of existing STGCN models, and enable us to realize the parameterized representation of STGCN models. However, if we only consider these parameters in the STGCN search space, then we may miss many powerful models. Specifically, we observe that structures of existing STGCN models, where ST-blocks are connected linearly and share the same structure (as is shown in Figure~\ref{fig2} (a)), are not flexible enough. Previous works on Neural Architecture Search (NAS)~\cite{p26,p27,p28} point out that block diversity and flexible connection method have great importance on model performance. This rule may also apply to our STGCN study, which focuses on similar neural networks as NAS. We may discover more effective STGCN models if we break the existing structural mode. Motivated by this, we introduce more parameters to consider more flexible and diverse STGCN structures. 

Suppose there are $N$ ST-block in a STGCN model, we allow each ST-block have different structural settings, so as to increase the structural diversity. That is to say we use totally $N$ groups of parameters ($SIPM^{b_{i}}$, $TIPM^{b_{i}}$, $FES^{b_{i}}$) ($i=1,\ldots,N$)\footnote{$SIPM^{b_{i}}$, $TIPM^{b_{i}}$, $FES^{b_{i}}$ represent the \textit{SIPM}, \textit{TIPM} and \textit{FES} of ${i}^{th}$ ($i=1,\ldots,N$) ST-block in the STGCM model, and have the same value spaces as $SIPM$, $TIPM$, $FES$ respectively.} to determine the structure of $N$ ST-blocks respectively. Besides, we design the following parameters to describe the flexible connection method among ST-blocks and the way of generating the final output of Stage2 in our Unified-STGCN utilizing these ST-blocks.

(1) Pre Block Index (\textit{PBIndex}): Different from previous STGCN models, we do not restrict to use the sequential connection method, but allow each ST-block connect with any one of its previous ST-blocks instead, and thus construct more flexible STGCN structures. We introduce parameters $PLIndex^{b_{i}}$ ($i=1,\ldots,N$) to clarify the inputs of ST-blocks in STGCN, and their options are as follows.
\begin{equation}
\setlength{\abovedisplayskip}{0.1cm}
\setlength{\belowdisplayskip}{-0.02cm}
PLIndex^{b_{i}}\in \{b_{1},\ldots,b_{i-1}\}
\end{equation}
Setting $PLIndex^{b_{i}}$ to $b_{j}$ means that taking the output of $j^{th}$ ST-block as the input of $i^{th}$ ST-block.

(2) Multiple ST-blocks Output Fusion method (\textit{MBOF}): The flexible connection method in our STGCN models may result in multiple output of Stage2, i.e., there may have multiple ST-blocks which are not considered as the preceding ST-block of any other ST-blocks. Under such circumstance, we aggregate multiple outputs to one using add or concentration approach, and we use \textit{MBOF} to describe this operation. 
\begin{itemize}
\item Add Aggregation Approach (denoted by $MBOF_{1}$): Adjust the shape of multiple outputs to be the same by using fully connected operation, then add them together. 
\item Concentration Aggregation Approach (denoted by $MBOF_{2}$): Adjust the feature dimension of multiple outputs to be the same by using fully connected operation, then concentrate them along the temporal dimension.
\end{itemize}

With the usage of these new parameters, we can obtain more flexible and diverse STGCN models (Figure~\ref{fig2} (b) is an example). Note that, filter size of convolution operations in ST-blocks is a hyperparameter of the STGCN model, which has great importance on model size. In this paper, we use parameter: Filter Size of Convolution (\textit{FSC})$\in\{16,32,64\}$ to control this hyperparameter, so as to obtain STGCN models with different sizes. Overall, we utilize $8+4\times N$ parameters to describe the design process of a STGCN model with $N$ ST-blocks. We then denote the configuration space of these parameters as $\mathbb{S}_{STGCN}$, where each configuration scheme $m\in \mathbb{S}_{STGCN}$ corresponds to a STGCN model.

\begin{table*}[t]
\caption{Four types of states that are used to describe the coding of a STGCN model. Considering a STGCN model which consists of n ST-blocks, (N+3) 5-D-vectors are used to describe its details, including its structure and its training setting.}
\vspace{-0.3cm}
\newcommand{\tabincell}[2]{\begin{tabular}{@{}#1@{}}#2\end{tabular}}
\centering
\resizebox{0.85\textwidth}{!}{
\smallskip\begin{tabular}{|m{1.9cm}<{\centering}|m{4.6cm}|m{1.1cm}<{\centering}|m{2.7cm}<{\centering}|m{2.7cm}<{\centering}|m{2.1cm}<{\centering}|m{2.9cm}<{\centering}|m{3.0cm}<{\centering}|}
\hline
\textbf{State Space Symbol} & \textbf{State Space Meaning} & \multicolumn{5}{c|}{\textbf{\tabincell{l}{State Space Description (5-D-vector): Contents and Their Settings}}} & \textbf{Action Space} \\
\hline
$State_{-2}$ & The start state of a Spatial-Temporal GCN model & \multicolumn{5}{c|}{[$-2,-1,-1,-1,-1$]} & \multirow{8}{*}{\tabincell{c}{\\ \\ \\ \\ \\ \\ \\ $Action_{i}=State_{i+1}$\\ \\ $i=\{-2,\ldots,N-1\}$}}\\
\cline{1-7}
\multirow{2}{*}{$State_{-1}$} & \multirow{2}{*}{\tabincell{l}{Set the details of the\\ $\bigcdot$ Stage4: Training Stage}} & \textbf{State Index} & \textbf{\textit{LF}} & \textbf{\textit{BS}} & \textbf{\textit{ILR}} & \textbf{\textit{OF}} & \\
\cline{3-7}
& & -1 & $LF_{1}$, $LF_{2}$ & $BS_{1}$,...,$BS_{3}$ & $ILR_{1}$,...,$ILR_{3}$ & $OF_{1}$,...,$OF_{3}$ & \\
\cline{1-7}
\multirow{2}{*}{\tabincell{c}{\\ \\ \\ $State_{0}$}} & \tabincell{l}{Set the details of the\\ $\bigcdot$ Stage1: Input Transform Stage} & \textbf{State Index} &\textbf{\textit{IS}} & \textbf{\textit{OS}} & \textbf{\textit{FI}} & \textbf{\textit{MBOF}} & \\
\cline{3-7}
& \tabincell{l}{$\bigcdot$ Stage3: Output Transform Stage\\Set \textit{FI} and \textit{MBOF} of the\\ $\bigcdot$ Stage2: Spatial-Temporal Embe-\\dding Stage} & \tabincell{c}{\\0} & \tabincell{c}{\\$IS_{1}$, $IS_{2}$} & \tabincell{c}{\\$OS_{1}$,...,$OS_{3}$} & \tabincell{c}{\\16, 32, 64} & \tabincell{c}{\\$MBOF_{1}$, $MBOF_{2}$} & \\
\cline{1-7}
\multirow{3}{*}{\tabincell{c}{\\ \\ $State_{i}$\\$i$ = $\{1,…,N\}$}} & \multirow{3}{*}{\tabincell{l}{\\ Set details of multiple ST-blocks\\ and their internal connections in\\ $\bigcdot$ Stage2: Spatial-Temporal Embe-\\dding Stage}} & \textbf{State Index} & \textbf{\textit{SIPM}} & \textbf{\textit{TIPM}} & \textbf{\textit{FES}} & \textbf{PBIndex} & \\
\cline{3-7}
& & $i$ & $SIPM_{1}$,...,$SIPM_{4}$ & $TIPM_{1}$,...,$TIPM_{3}$ & $FES_{1}$,...,$FES_{4}$ & $\{$1,2,...,$i-1\}$ $\cup$ $\{0\}$ Layer index 0 represents the output of Stage2 & \\
\cline{3-7}
& & \multicolumn{5}{c|}{Terminal State: [$i,-1,-1,-1,-1$]} & \\
\hline
\end{tabular}
}
\label{table2}
\vspace{-0.55cm}
\end{table*}

\section{Auto-STGCN: Autonomous STGCN Search}\label{section:4}

In this section, we propose Auto-STGCN to optimize parameters provided by Section~\ref{section:3}, and design the optimal STGCN model automatically. Section~\ref{section:4.1} gives our optimization target and Section~\ref{section:4.2} introduces Auto-STGCN in detail.

\subsection{Constraint-Aware Objective Function}\label{section:4.1}

Given a STGCN model $m\in \mathbb{S}_{STGCN}$, let $MAE(m)$ denote its Mean Absolute Errors score on the target Spatial-Temporal NDF task, $T(m)$ denote the inference time, and $T_{max}$ is the maximum time constraint. Formally, our research target is defined as follows.
\begin{equation}
\setlength{\abovedisplayskip}{0.1cm}
\setlength{\belowdisplayskip}{-0.02cm}
\min_{m\in \mathbb{S}_{STGCN}} MAE(m)\ \ subject\  to\ T(m)\leq T_{max}
\label{equ5}
\end{equation}
We hope to discover effective STGCN models whose prediction speed is not too slow. Therefore, we set $T_{max}$ to twice the inference time of \cite{p15}, a state-of-the-art STGCN model, on the target task. Note that, reinforcement learning fails to deal with such constraint. To guide reinforcement learning approach to discover STGCN models that satisfy the requirements, we construct a log barrier function to quantify the time constraint in Equation~(\ref{equ5}), and define a new constraint-aware objective function as follows. 
\begin{equation}
\setlength{\abovedisplayskip}{0.1cm}
\setlength{\belowdisplayskip}{-0.02cm}
\begin{split}
\min_{m\in \mathbb{S}_{STGCN}} MAE(m)-\lambda log(\frac{T_{max}}{T(m)})
\end{split}
\label{equ6}
\end{equation}
where $\lambda$ is set to $e^{-19}$, a very small value to make the constraint tight. As we can see, if $T(m)\leq T_{max}$, barrier function is close to being violated, whereas, the value of log barrier function approaches infinity when the constraint is violated. Equation~(\ref{equ6}) is a smooth approximation of the Equation~(\ref{equ5}), and reasonably describes our optimization target by using only one function. Later, we will apply this constraint-aware objective function to our Auto-STGCN, thus search powerful STGCN models which satisfy the time constraint.

\subsection{Auto-STGCN Algorithm}\label{section:4.2}

In \underline{Auto}nomous \underline{STGCN} Search (Auto-STGCN) algorithm, we employ the well-known Q-learning~\cite{p29}, a value-based reinforcement learning algorithm which uses Q function to find the optimal action-selection policy, with epsilon-greedy strategy~\cite{p23} to effectively and automatically search the optimal STGCN model. State and actions in Auto-STGCN are defined as follows. 

We utilize $N+3$ states $(s_{-2},s_{-1},\ldots, s_{N})$ $s_{i}\in State_{i}$ (as is shown in Table~\ref{table2}) to describe the setting of $8+4\times N$ parameters designed in Section~\ref{section:3.2}, thus completely and clearly describe a STGCN model. Specifically, $s_{-2}$=[-2,-1,-1,-1,-1] is an initial state; $s_{-1}$ $\in$ $State_{-1}$=$\{$-1, \textit{LF}, \textit{BS}, \textit{ILR}, \textit{OF}$\}$ shows the training details of the STGCN model; $s_{0}$ $\in$ $State_{0}$=$\{$0, \textit{IS}, \textit{OS}, \textit{FSC}, \textit{MBOF}$\}$ determines the input structure and output structure applied in STGCN, and sets the filter size and the method of dealing with multiple outputs of ST-blocks in STGCN model; $s_{i}$ $\in$ $State_{i}=$$\{i$, \textit{SIPM}, \textit{TIPM}, \textit{FES}, \textit{PBIndex}$\}$ $\cup$ $\{$[$i$,-1,-1,-1,-1]$\}$ ($i=1,\ldots,N$)\footnote{The [$i$,-1,-1,-1,-1] ($i=1,$$\ldots$$,N$) are terminal states. The $s_{i+1}$,$\ldots$,$s_{N}$ becomes invalid when $s_{i}$ is set to [$i$,-1,-1,-1,-1].} elaborates on the connection details and structure details of ST-blocks in STGCN. For each state $s_{i}\in State_{i}$ ($i=-2,$$\ldots,$$N-1$), we define its action space as $Action_{i}=State_{i+1}$ , and use $A(s_{i})$$\in$$Action_{i}$ to decide for its next successive state. Accordingly, the design process of a STGCN model can be considered as an action selection trajectory, and our state transition process is described as follows.
\begin{equation}
\setlength{\abovedisplayskip}{0.1cm}
\setlength{\belowdisplayskip}{-0.02cm}
\begin{split}
s_{-2} = [&-2,-1,-1,-1,-1]\\
&(s_{i},A(s_{i}))\rightarrow s_{i+1}\\  
A(s_{i})\in Action_{i},\ &s_{i}\in State_{i},\ i=-2,…N-1
\end{split}
\end{equation}

We model the action selection process as a Markov Decision Process~\cite{p30}. In order to find the optimal STGCN model, we ask the agent to maximize its expected reward over all possible trajectories, and utilize recursive Bellman Equation~\cite{p32} to deal with this maximization problem. Given a state $s_{i}\in State_{i}$, and subsequent action $A(s_{i})\in Action_{i}$, we denote the maximum expected accumulative reward that the agent would receive as $Q^{*}(s_{i},A(s_{i}))$, and the recursive Bellman Equation can be written as follows.
\begin{equation}
\scriptsize
\setlength{\abovedisplayskip}{0.1cm}
\setlength{\belowdisplayskip}{-0.02cm}
\begin{split}
Q^{*}(s_{i},A(s_{i}))=&\mathbb{E}_{s_{i+1} |s_{i},A(s_{i})} [\mathbb{E}_{r|s_{i},A(s_{i}),s_{i+1}} [r|s_{i},A(s_{i}),s_{i+1}]\\
&+\gamma \max_{a\in Action_{i+1}} Q^{*}(s_{i+1},a)]
\end{split}
\end{equation}
where $\gamma$ is the discount factor which measures the importance of the future rewards. Formulating the above equation as an iterative update, then we get the following equations:
\begin{equation}
\scriptsize
\setlength{\abovedisplayskip}{0.1cm}
\setlength{\belowdisplayskip}{-0.02cm}
\begin{split}
Q(s_{T},None)&=0\\
Q(s_{T-1},A(s_{T-1}))&=(1-\alpha) Q(s_{T-1},A(s_{T-1}))+\alpha r_{T-1}\\
Q(s_{i},A(s_{i}))&=(1-\alpha) Q(s_{i},A(s_{i}))+\alpha [r_{i}\\
+\gamma \max_{a\in Action_{i+1}}& Q(s_{i+1},a)], i\in \{-2,\ldots,T-2\}
\end{split}
\end{equation}
where $\alpha$ is the learning rate which determines how the newly acquired information overrides the old Q-value, $r_{i}$ ($i=-2,\ldots,T-1$) denotes the intermediate reward observed for the current state $s_{i}$ after taking action $A(s_{i})$, and $s_{T}$ $\in$ $\{[i$,-1,-1,-1,-1]$|i$=1,…,$N-1\}$$\cup$ $State_{N}$ refers to a terminal state. Note that rewards: $r_{-2},\ldots,r_{T-2}$ cannot be explicitly measured in our task. Ignoring them in the iterative process by setting them to 0, may causes a slow convergence in the beginning~\cite{p33} and thus makes Auto-STGCN time consuming. To speed up the learning process of agent, we introduce reward shaping~\cite{p31} method and apply the following shaped intermediate reward instead in our Auto-STGCN algorithm\footnote{This method has been proved to be able to accelerate the learning process of agent in \cite{p33}}.
\begin{equation}
\setlength{\abovedisplayskip}{0.1cm}
\setlength{\belowdisplayskip}{-0.02cm}
\begin{split}
r_{i}=\frac{R_{s_{-2}\sim s_{T}}}{T+1}=&\frac{-MAE(m)+\lambda log(\frac{T_{max}}{T(m)})}{T-1}\\ 
\end{split}
\end{equation}
where $R_{s_{1}\sim s_{T}}$ is the validation performance of corresponding STGCN model trained convergence on training set for the trajectory $(s_{1},\ldots,s_{T})$. In Auto-STGCN, we utilize Q-value combined with epsilon-greedy strategy to select the STGCN model to be evaluated for each iteration, and update Q-value according to the evaluation information and Equation (9). We provide the pseudo-code and convergence proof of Auto-STGCN in the supplementary material.

\section{Experiments}\label{section:5}

In this section, we evaluate the Auto-STGCN algorithm. We implement all experiments using MXNet~\cite{p34}.

\subsection{Experimental Setting}\label{section:5.1}

\textbf{Datasets.} In the experiment, we use four high-way traffic datasets: PEMS03, PEMS04, PEMS07 and PEMS08, which are collected by the Caltrans Performance Measurement System (PeMS)~\cite{p35}. For each dataset, we split all datasets with ratio 6:2:2 into training sets, validation sets and test sets. We use the past 12 continuous time steps to predict the future 12 continuous time steps. 

\begin{table}[t]
\caption{Performance comparison of different STGCN models. Time refers to the Inference time on test sets.}
\vspace{-0.3cm}
\newcommand{\tabincell}[2]{\begin{tabular}{@{}#1@{}}#2\end{tabular}}
\centering
\resizebox{1.0\columnwidth}{!}{
\smallskip\begin{tabular}{m{1.1cm}<{\centering}|m{1.0cm}<{\centering}|m{1.6cm}<{\centering}|m{1.6cm}<{\centering}|m{1.9cm}<{\centering}|m{1.6cm}<{\centering}|m{1.9cm}<{\centering}}
\hline
\multirow{2}{*}{\textbf{Datasets}} & \multirow{2}{*}{\textbf{Metrics}} & \multicolumn{5}{c}{\textbf{Algorithm}} \\
\cline{3-7}
& & \textbf{STGCRN} & \textbf{ASTGCN} & \textbf{STGCN(2018)} & \textbf{STSGCN} & \textbf{AutoSTGCNM} \\
\hline
\multirow{3}{*}{PEMS03} & MAE & 24.23$\pm$1.02 & 17.69$\pm$1.43 & 17.49$\pm$0.46 & 17.48$\pm$0.15 & \textbf{16.43$\pm$0.11} \\
 & MAPE & 21.44$\pm$0.62 & 19.40$\pm$2.24 & 17.15$\pm$0.46 & 16.78$\pm$0.20 & \textbf{15.33$\pm$0.41} \\
 & RMSE & 47.71$\pm$3.11 & 29.66$\pm$1.68 & 30.12$\pm$0.70 & 29.21$\pm$0.56 & \textbf{25.65$\pm$0.17} \\
\hline
\multirow{3}{*}{PEMS04} & MAE & 27.27$\pm$2.15 & 22.93$\pm$1.29 & 22.70$\pm$0.64 & 21.19$\pm$0.10 & \textbf{20.63$\pm$0.22} \\
 & MAPE & 17.31$\pm$1.12 & 16.56$\pm$1.36 & 14.50$\pm$0.21 & 13.90$\pm$0.05 & \textbf{13.69$\pm$0.30} \\
 & RMSE & 50.88$\pm$7.29 & 35.22$\pm$1.90 & 35.55$\pm$0.75 & 33.65$\pm$0.20 & \textbf{31.30$\pm$0.32} \\
\hline
\multirow{3}{*}{PEMS07} & MAE & 52.01$\pm$5.66 & 28.05$\pm$2.34 & 25.38$\pm$0.49 & 24.26$\pm$0.14 & \textbf{23.47$\pm$0.12} \\
 & MAPE & 22.61$\pm$2.65 & 13.92$\pm$1.65 & 11.08$\pm$0.18 & 10.21$\pm$0.05 & \textbf{10.09$\pm$0.15} \\
 & RMSE & 79.34$\pm$4.59 & 42.57$\pm$3.31 & 38.78$\pm$0.58 & 39.03$\pm$0.27 & \textbf{35.97$\pm$0.17} \\
\hline
\multirow{3}{*}{PEMS08} & MAE & 23.02$\pm$0.77 & 18.61$\pm$0.40 & 18.02$\pm$0.14 & \textbf{17.13$\pm$0.09} & 17.16$\pm$0.31 \\
 & MAPE & 13.83$\pm$0.36 & 13.08$\pm$1.00 & 11.40$\pm$0.10 & 10.96$\pm$0.07 & \textbf{10.60$\pm$0.19} \\
 & RMSE & 37.66$\pm$3.39 & 28.16$\pm$0.48 & 27.83$\pm$0.20 & 26.80$\pm$0.18 & \textbf{25.84$\pm$0.44} \\
\hline
\end{tabular}
}
\label{table3}
\vspace{-0.5cm}
\end{table}

\begin{figure}[t]
\begin{minipage}[t]{0.45\columnwidth}
\centering
\includegraphics[height=3.7cm,width=4.0cm]{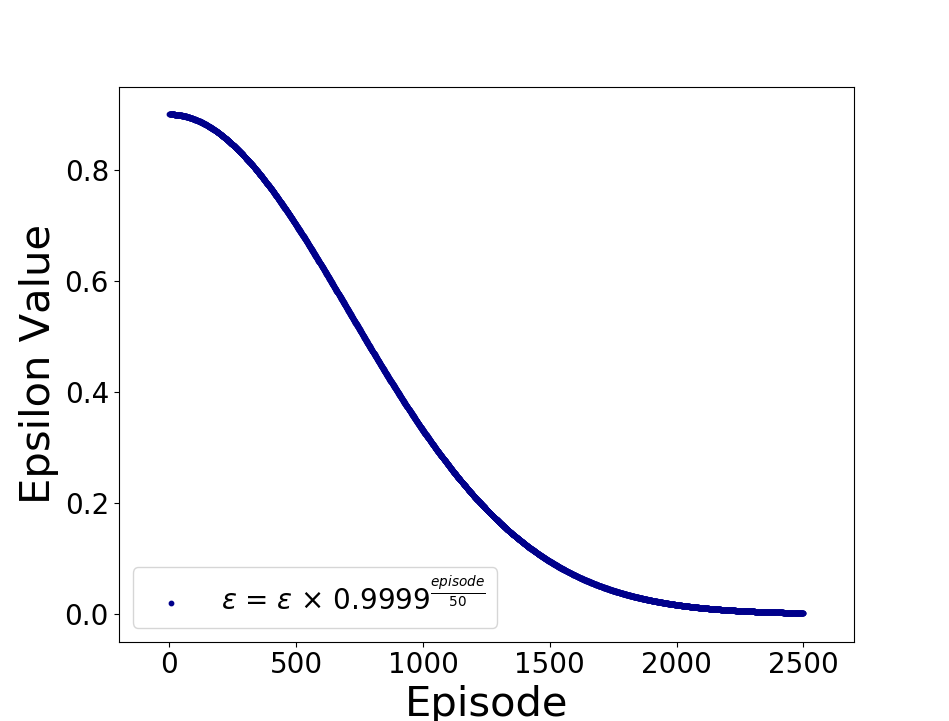}
\vspace{-0.7cm}
\caption{\small{Epsilon values applied in Auto-STGCN.}}
\label{fig3}
\end{minipage}
\begin{minipage}[t]{0.54\columnwidth}
\centering
\includegraphics[height=5.1cm,width=3.9cm]{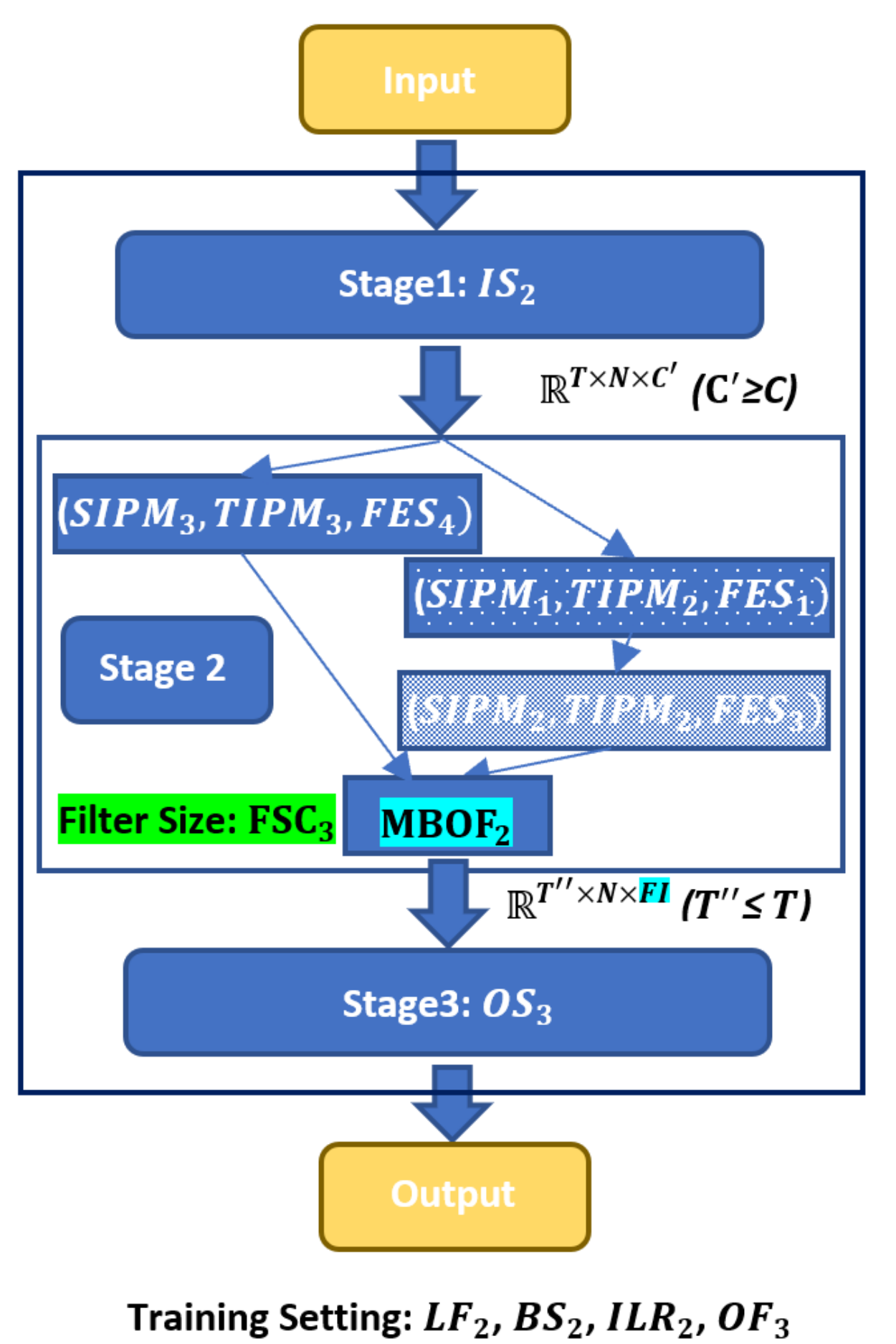}
\vspace{-0.3cm}
\caption{\small{Details of the \textit{AutoSTGCNM} searched by Auto-STGCN.}}
\label{fig4}
\end{minipage}
\vspace{-0.3cm}
\end{figure}

\textbf{Implementation details of Auto-STGCN.} The maximum number of ST-blocks $N$ is set to 4. In the Q-value update process, learning rate $\alpha$ is set to 0.001, and discount factor $\gamma$ is set to 0.9. During the searching phase, we train the agent with 2000 episodes, i.e., sampling 2000 STGCN models in total. For each generated STGCN model, we train it for a fixed 5 epochs on PEMS03 dataset, and measure its performance according to its MAE score and inference latency on validation sets. As for the epsilon-greedy strategy applied in Auto-STGCN, we decrease $\varepsilon$ from 0.9 to 0.0 following the epsilon schedule as shown in Figure~\ref{fig3}. 

Our Auto-STGCN takes about 4.75 GPU days to accomplish the search phase on a single NVIDIA Tesla V100 GPU. After obtaining the best auto-generated STGCN model searched on PEMS03, which is denoted by \textit{AutoSTGCNM}, we train it for 50 epochs on PEMS03 dataset, and report its performance scores on the test set. We also evaluate the transfer ability of \textit{AutoSTGCNM} to the other 3 datasets, i.e., PEMS04, PEMS07 and PEMS08. Figure~\ref{fig4} shows the details of \textit{AutoSTGCNM}.

\subsection{Effectiveness of Auto-STGCN}\label{section:5.2}

In this part, we examine the effectiveness of Auto-STGCN. We compare \textit{AutoSTGCNM} with 4 state-of-the-art STGCN models discussed in Section~\ref{section:3.1}: STSGCN~\cite{p15}, ASTGCN~\cite{p16}, STGCN(2018)~\cite{p14}, STGCRN~\cite{p17}, using four Spatial-Temporal NDF tasks. Results are shown in Table~\ref{table3}. Our \textit{AutoSTGCNM} consistently outperforms existing STGCN methods on three datasets except for PEMS08. In PEMS08, \textit{AutoSTGCNM} has the best MAPE and RMSE, except for MAE which is slightly larger than that of STSGCN. Taking operations of existing STGCN models as components, Auto-STGCN designs a more powerful STGCN model by integrating advantages of different models, which demonstrates the effectiveness of our approach.

\subsection{Importance of Diversity and Flexibility}\label{section:5.3}

We further investigate the effect of diverse ST-block structures and flexible connection method on the performance of STGCN models using the following three variants of \textit{AutoSTGCNM}, thus examine the reasonability of search space designed in our Auto-STGCN. 
\begin{itemize}
\item[1.] \textit{-Diversity}: This model changes the structures of all ST-blocks in \textit{AutoSTGCNM} model to be the same as that of the third ST-block in \textit{AutoSTGCNM}, i.e., ($SIPM_{2}$,$TIPM_{2}$,$FES_{3}$).
\item[2.] \textit{-Connection Flexibility}: This model changes the connection method of ST-blocks in \textit{AutoSTGCNM} model to ST-block1$\rightarrow$ST-block2$\rightarrow$ST-block3.
\item[3.] \textit{-Multiple Source}: This model changes the structures of all ST-blocks in \textit{-Connection Flexibility} model to ($SIPM_{3}$,$TIPM_{2}$,$FES_{4}$), where all operations related to ST-blocks come from the same paper, i.e., STSGCN~\cite{p15}.
\end{itemize}

\begin{figure}[t]
\centering
\includegraphics[width=0.95\columnwidth]{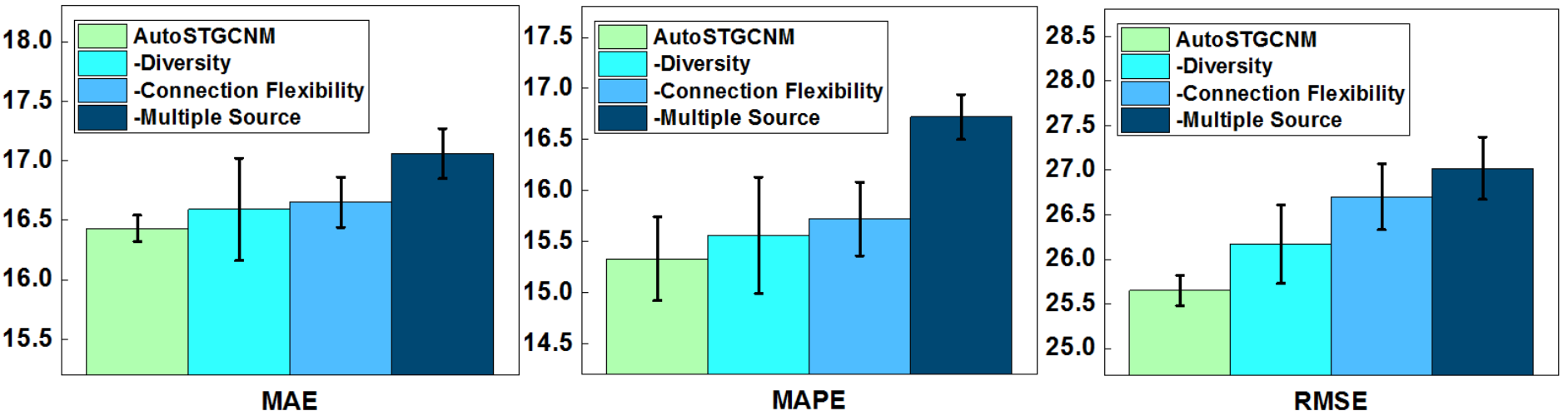}
\vspace{-0.3cm}
\caption{Performance of three variants of \textit{AutoSTGCNM} on PEMS03 dataset.}
\label{fig5}
\vspace{-0.5cm}
\end{figure}

As Figure~\ref{fig5} illustrates, \textit{AutoSTGCNM} has much better performance than \textit{-Diversity} and \textit{-Connection Flexibility}. This result shows us that applying diverse ST-block structures and the flexible connection method among ST-blocks can effectively improve the performance of the STGCN model, which coincides with our discussion in Section~\ref{section:3.2} and demonstrates the reasonability of search space designed in our Auto-STGCN. Besides, we observe that \textit{-Multiple Source} performs the worst among them. This result shows us that the significance and necessity of breaking the original combinations of existing STGCN models. More powerful STGCN models can be found by combining excellent operations of different STGCN models, which demonstrates the reasonability of our approach.

\section{Conclusion and Future Works}\label{section:6}

In this paper, we propose Auto-STGCN to help users automatically design high-performance STGCN models using existing work, and thus effectively deal with practical Spatial-Temporal NDF problems. Our approach breaks the original combinations making excellent operations of different STGCN models capable of being combined together, and discovers more powerful STGCN models by integrating advantages of multiple models and applying effective STGCN optimization method. Extensive experiments on real-world benchmark datasets show that our Auto-STGCN can find STGCN models superior to existing STGCN models with heuristic parameters, which demonstrates the effectiveness of our proposed method. In the future work, we will try to propose effective methods to quickly evaluate candidates STGCN models analyzed in the optimization process to further accelerate the optimization speed.

\end{document}